\documentclass[lettersize,journal]{IEEEtran}
\usepackage{amsmath,amsfonts}
\usepackage{algorithmic}
\usepackage{algorithm}
\usepackage{array}
\usepackage[caption=false,font=normalsize,labelfont=sf,textfont=sf]{subfig}
\usepackage{textcomp}
\usepackage{stfloats}
\usepackage{url}
\usepackage{verbatim}
\usepackage{graphicx}
\usepackage{cite}
\usepackage{times}
\usepackage{epsfig}
\usepackage{graphicx}
\usepackage{amssymb}
\usepackage{mathabx}
\usepackage{pifont}
\usepackage{fontawesome}
\usepackage{color}
\begin{document}

\title{MBMamba: When Memory Buffer Meets Mamba for Structure-Aware Image Deblurring}

\author{Hu Gao, Xiaoning Lei, Xichen Xu, Depeng Dang, Lizhuang Ma$^\dag$
        % <-this % stops a space
\thanks{Hu Gao, Xichen Xu, and Lizhuang Ma are with Shanghai Jiao Tong University. (e-mail: gao\_h@sjtu.edu.cn).

$\dag$ Corresponding Author
}}% <-this % stops a space

\IEEEpubid{\begin{minipage}{\textwidth}\  \\[30pt] \centering
		Copyright © 20xx IEEE. Personal use of this material is permitted. However, permission to use this material for any other purposes must be obtained from the IEEE by sending an email to pubs-permissions@ieee.org.
\end{minipage}}

\maketitle

\begin{abstract}
The Mamba architecture has emerged as a promising alternative to CNNs and Transformers for image deblurring. However, its flatten-and-scan strategy often results in local pixel forgetting and channel redundancy, limiting its ability to effectively aggregate 2D spatial information. Although existing methods mitigate this by modifying the scan strategy or incorporating local feature modules, it increase computational complexity and hinder real-time performance. In this paper, we propose a structure-aware image deblurring network  without changing the original Mamba architecture. Specifically, we design a memory buffer mechanism to preserve historical information for later fusion, enabling reliable modeling of relevance between adjacent features. Additionally, we introduce an Ising-inspired regularization loss that simulates the energy minimization of the physical system's "mutual attraction" between pixels, helping to maintain image structure and coherence. Building on this, we develop MBMamba. Experimental results show that our method outperforms state-of-the-art approaches on widely used benchmarks.

\end{abstract}


\section{Introduction}
Image deblurring seeks to restore a sharp latent image from a blurred observation. Given the ill-posed nature of this inverse problem, traditional methods~\cite{karaali2017edge, 2011Image} often incorporate handcrafted features or explicit priors to narrow the solution space toward natural images. However, designing such priors is not only challenging but also lacks generalizability, making them less effective in real-world applications.

Benefiting from the rapid progress of deep learning in high-level vision tasks, numerous data-driven methods have adopted CNNs as backbone architectures~\cite{10937503,FSNet}. Although convolutions are effective at capturing local patterns, their inherent limitations—such as restricted receptive fields and content-agnostic operations—hinder their ability to model long-range dependencies. To address these issues, several approaches~\cite{xu2025motion,kong2023efficient,u2former,Zamir2021Restormer} have introduced transformers into image deblurring, achieving superior performance over CNN-based methods by leveraging attention mechanisms within spatial windows or across channel dimensions. Nonetheless, these methods still face challenges: they either struggle to fully utilize spatial details or are constrained by coarse partitioning strategies that limit the extraction of fine-grained features within individual windows.

\begin{figure}
    \centering
    \includegraphics[width=1\linewidth]{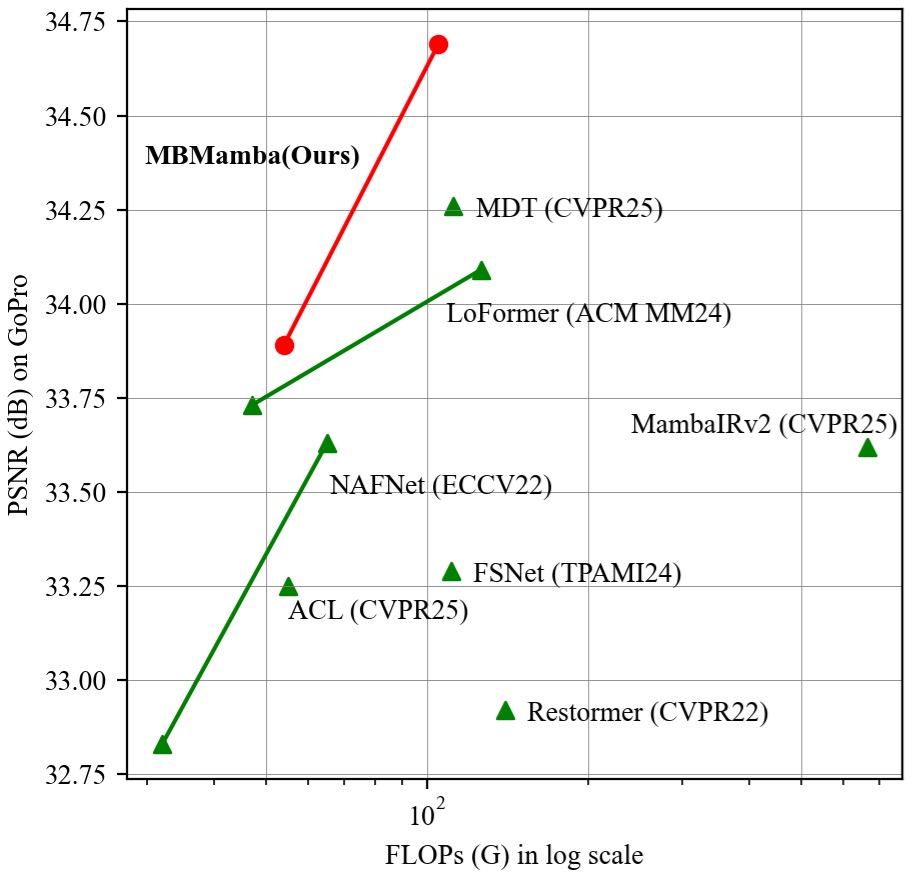}
    \caption{Computational cost vs. PSNR of models on the GoPro dataset~\cite{Gopro}. Our MBMamba achieve the SOTA performance while simultaneously reducing computational costs.}
    \label{fig:param}
\end{figure}

State space models~\cite{gu2023mamba,ssmehta2022long}, especially the enhanced Mamba variant, have recently attracted considerable attention for their ability to model long-range dependencies with linear complexity. However, the widely adopted flatten-and-scan strategy often leads to the loss of local pixel information and redundant channel features, limiting the model’s capacity to effectively capture 2D spatial structures. Given the importance of local detail and channel-wise cues in image deblurring, directly applying state space models typically results in subpar performance. To overcome these limitations, recent studies have proposed a variety of scanning strategies—such as four-directional scanning~\cite{guo2024mambair}, shortest path traversal~\cite{Zhou_2025_CVPR}, and slice-and-scan~\cite{liu2024xyscannet}—as well as the integration of local feature enhancement modules. However, these solutions often require multiple scans or additional components, which inevitably increase computational overhead and compromise real-time efficiency.

Based on the above analysis, a natural question arises: can we design a structure-aware image deblurring network  that efficiently integrates both local and global features without increasing the number of scans or adding extra local modules? To address this, we propose MBMamba, which incorporates several key components.
Specifically, we design a memory buffering mechanism that stores historical information, which is then fused with the current features via a cross-attention mechanism, enabling more robust modeling of dependencies between adjacent features. 
In addition, we present an Ising-inspired regularization loss that mimics the energy minimization process of physical systems, capturing the "mutual attraction" between pixels to better preserve image structure and coherence.
Furthermore, to fully exploit the decoder’s potential, we utilize a pre-trained encoder and implement multiple sub-decoders alongside multi-scale output designs, thereby easing the training process.
As demonstrated in Figure~\ref{fig:param}, MBMamba achieves state-of-the-art results while maintaining computational efficiency compared to existing approaches.

The main contributions of this work are:
\begin{enumerate}
	\item We propose MBMamba, a structure-aware image deblurring network that efficiently integrates both local and global features.
    
    \item We design a memory buffer mechanism that preserves historical information for subsequent fusion, enabling reliable modeling of relationships between adjacent features.
    
    \item We present an Ising-inspired regularization loss to capture the "mutual attraction" between pixels to better maintain image structure and coherence.
    
    \item Extensive experiments show that MBMamba achieves competitive performance compared to state-of-the-art methods.
\end{enumerate}

\section{Related Work}
\subsection{Hand-crafted prior-based methods}
Given the inherently ill-posed nature of image deblurring, early methods~\cite{karaali2017edge, tra1910.1007/978-3-030-58595-2_38, tra209241002} often relied on manually designed priors to narrow the solution space. Although some approaches have attempted to incorporate additional sensor data, such as inertial measurements, to estimate blur kernels more accurately~\cite{trannnnn10558778}, these prior-driven strategies generally struggle to model the complex degradation process and often lack robustness and general applicability. Moreover, designing effective priors typically involves complex optimization procedures, limiting their practical use.

\subsection{CNN-based methods}
With the rapid progress of deep learning, many approaches have shifted from manually crafting image priors to developing various CNN-based models for image deblurring. To effectively balance spatial detail preservation and contextual understanding, MPRNet~\cite{Zamir2021MPRNet} introduces cross-stage feature fusion to utilize features from multiple processing stages.
MIRNet-V2~\cite{Zamir2022MIRNetv2} adopts a multi-scale design to extract richer features for restoration tasks, while IRNeXt~\cite{IRNeXt} reconsiders CNN architecture to build a more efficient and effective network.
NAFNet~\cite{chen2022simple} simplifies the model structure by analyzing and refining baseline components, removing or replacing non-linear activations. SFNet~\cite{SFNet} and FSNet~\cite{FSNet} propose dynamic and compact frequency selection modules that identify the most informative components for restoration.
ELEDNet~\cite{elednetkim2024towards} combines cross-modal information with low-pass filtering to suppress noise while retaining structure. MR-VNet~\cite{MR-VNet} leverages Volterra layers for efficient blur removal. DSDNe~\cite{10937503} formulates the deblurring task as separate data and regularization sub-problems to improve speed and accuracy.
While these CNN-based methods outperform traditional prior-driven approaches, their reliance on local receptive fields inherently limits their ability to address long-range degradation patterns effectively.

\subsection{Transformer-based methods}
Thanks to their content-aware global receptive fields, Transformer architectures~\cite{2017Attention} have recently become increasingly popular in image restoration, consistently outperforming traditional CNN-based methods.
However, image deblurring often involves high-resolution inputs, where the quadratic computational complexity of standard attention mechanisms leads to heavy processing overhead. To mitigate this, models like Uformer~\cite{Wang_2022_CVPR}, SwinIR~\cite{liang2021swinir}, and U$^2$former~\cite{u2former} adopt window-based self-attention to localize computation. Yet, this strategy limits the capacity to capture full contextual information within each patch.
To improve efficiency, Restormer~\cite{Zamir2021Restormer} and MRLPFNet~\cite{MRLPFNet} shift attention computation to the channel dimension, achieving linear complexity. Nevertheless, this design compromises spatial feature modeling.
FFTformer~\cite{kong2023efficient} explores attention computation in the frequency domain using Fourier transforms, but requires inverse operations that introduce additional cost.
MAT~\cite{xu2025motion} proposes a motion-adaptive Transformer, leveraging motion cues to build more robust global dependencies.
For more realistic deblurring scenarios, HI-Diff~\cite{diffdeblurNEURIPS2023_5cebc89b} incorporates diffusion models to generate informative priors, which are then hierarchically integrated to enhance the deblurring process.

\begin{figure*}
    \centering
    \includegraphics[width=1\linewidth]{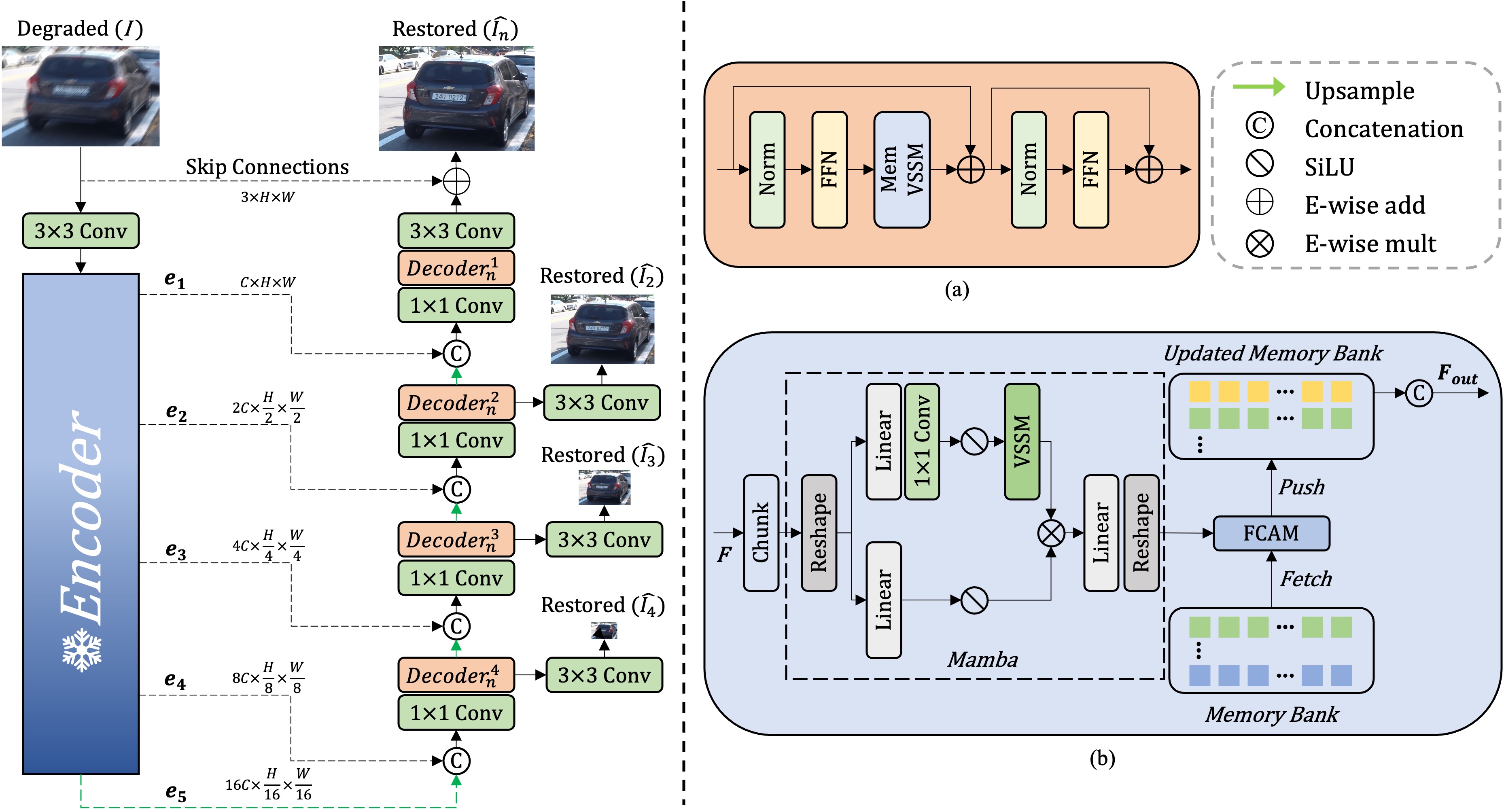}
    \caption{The overall architecture of the proposed MBMamba: (a) The decoder is composed of vision state space models equipped with a memory buffering mechanism (MemVSSM); (b) The internal structure of MemVSSM.}
    \label{fig:network}
\end{figure*}

\subsection{State Spaces Model}
State space models~\cite{s4gu2021combining,s5smith2022simplified} have recently gained prominence for their efficiency in modeling long-range dependencies with linear computational complexity.
Mamba~\cite{gu2023mamba} enhances this framework with a selective mechanism and a hardware-friendly parallel algorithm. Nonetheless, when applied to image restoration tasks, the standard Mamba still suffers from issues such as local pixel information loss and redundant channel features.
To overcome these limitations, several recent approaches have explored alternative scanning strategies and incorporated modules aimed at enhancing local feature representation. MambaIR~\cite{guo2024mambair} introduces a four-directional unfolding method combined with channel attention. ALGNet~\cite{algnetgao2024learning} leverages a fusion module that jointly captures global and local information for more accurate feature extraction.
LoFormer~\cite{xintm2024LoFormer} applies local channel-wise self-attention in the frequency domain to model cross-covariance within low- and high-frequency local regions. XYScanNet~\cite{liu2024xyscannet} proposes an alternating slice-and-scan strategy along intra- and inter-slice directions.
MambaIRV2~\cite{guo2025mambairv2} further enhances Mamba by introducing non-causal modeling similar to ViTs, enabling a more attentive state space representation.

However, these enhancements typically come at the cost of increased complexity, requiring additional scanning passes or supplementary modules, which hinders real-time performance.
To address these challenges, we propose MBMamba to effectively integrates both local and global features for image deblurring—without incurring the overhead of multiple scans or added modules.

\section{Method}

In this section, we first provide an overview of the entire MBMamba pipeline. We then dive into the details of the proposed decoder, which comprises vision state space models equipped with a memory buffering mechanism (MemVSSM). Lastly, we present the Ising-inspired regularization loss. 

\subsection{Overall Pipeline} 
Our proposed MBMamba, illustrated in Figure~\ref{fig:network}, consists of a frozen encoder and $n$ sub-decoders, each comprising four decoding stages. Given a degraded image $\mathbf{I} \in \mathbb{R}^{H \times W \times 3}$, MBMamba first uses a convolutional layer to extract shallow features $\mathbf{F} \in \mathbb{R}^{H \times W \times C}$, where $H$, $W$, and $C$ denote the height, width, and number of channels of the feature map, respectively. These shallow features are passed through a pre-trained encoder to obtain multi-scale encoder features $e_i$ (where $i = 1, 2, 3, 4, 5$) at different scales. The encoder features are then fed into the decoder, which generates decoder features $d^i_n$ at different scales, progressively restoring them to their original size. Notably, since MBMamba incorporates multiple sub-decoders, the input to each subsequent sub-decoder is the output of the previous one. Finally, a convolutional layer is applied to the refined features to generate the residual image $\mathbf{X_n} \in \mathbb{R}^{H \times W \times 3}$ for $n_{th}$ sub-decoder. This residual image is added to the degraded image to produce the restored output: $\mathbf{\hat{I}_n} = \mathbf{X_n} + \mathbf{I}$.

\subsection{MemVSSM} 
The Mamba architecture has recently emerged as a promising alternative to CNNs and Transformers for image deblurring. However, its flatten-and-scan strategy tends to cause local pixel forgetting and channel redundancy, which undermines its ability to effectively capture 2D spatial information. While some existing approaches address this issue by altering the scanning strategy or adding local feature modules, these modifications often lead to increased computational cost and reduced real-time performance. To better capture both local and global features without increasing scan frequency or introducing additional modules, we propose a vision state space models equipped with a memory buffering mechanism (MemVSSM)  in the decoder. 
As shown in Figure~\ref{fig:network}(a), given the input features at the $(l-1)_{th}$ block $X_{l-1}$, the procedures of decoder can be defined as:
\begin{equation}
\begin{aligned}
\label{eq:dh1}
    X_l^{'} &= MemVSSM(FFN(Norm(X_{l-1}))) \oplus X_{l-1}
    \\
    X_l &= FFN(Norm(X_l^{'})) \oplus  X_l^{'}
\end{aligned}
\end{equation}

Previous Mamba-based methods~\cite{liu2024xyscannet,guo2025mambairv2} input image features into the VSSM at once, which necessitates additional scan iterations or the introduction of local feature modules to mitigate local pixel forgetting and channel redundancy caused by an excessive number of hidden units.

In contrast, as illustrated in Figure~\ref{fig:network}(b), our MemVSSM first divides the input feature $F$ into $N$ channel-wise chunks.
\begin{equation}
\label{eq:mem1}
    F^d_1,..., F^d_i,...,F^d_N  = Chunk(F)
\end{equation}

Each chunk is sequentially processed by Mamba to produce enriched feature representations $F_i$, where $i$ denotes the chunk index. A memory bank is then introduced to retain historical information for subsequent fusion. Specifically, we define the memory bank with the depth of $K$, which indicates the number of stored historical feature sets.
The memory bank operates under a first-in, first-out (FIFO) policy for feature updating. In detail, during the processing of the $i$-th chunk, we first retrieve $K$ previous outputs [$F_{i-1}', F_{i-2}', ..., F_{i-K}'$], and fuse them with the current feature $F_i$ to obtain an enhanced representation $F_i'$ enriched with local contextual information. Then, the current feature $F_i'$ is pushed into the memory bank, and the oldest entry $F_{i-K}'$ is removed to maintain the buffer size.  Finally, we concatenate all the features in memory bank to get the final output feature $F_{out}$. Formally, each divided feature segment $F^d_i$ is first passed through Mamba to produce the output $F_i$, as follows:
\begin{equation}
\begin{aligned}
\label{equ:global}
    F_i &= Mamba(F^d_i) = Reshape(Linear(F^t_i \otimes  F^b_i))
    \\
    F^t_i & = VSSM(SiLU(f_{1 \times 1}^c(Linear(Reshape(F^d_i)))
    \\
    F^b_i &= SiLU(Linear(Reshape(F^d_i)))
\end{aligned}
\end{equation}
where $f_{1 \times 1}^c$ represents $1 \times 1$ convolution.

Then we fuse the current feature $F_i$ with the historical information [$F_{i-1}', F_{i-2}', ..., F_{i-K}'$] in the memory bank to obtain the feature  $F_i'$  with enhanced local information as follows:
\begin{equation}
\begin{aligned}
\label{equ:mem3}
    &F_i' = FCAM(F_i, F_{i-1}', F_{i-2}', ..., F_{i-K}')
    \\
    &F_{i-1}', F_{i-2}', ..., F_{i-K}'  = fetch(Memory Bank)
    \\
    & Updates Memory Bank = POP(F_{i-K}') \& PUSH(F_i')
\end{aligned}
\end{equation}
where $FCAM(\cdot)$ denotes the feature cross-attention mechanism, which enhances the current features by integrating information from the memory bank. 

\begin{figure}
    \centering
    \includegraphics[width=1\linewidth]{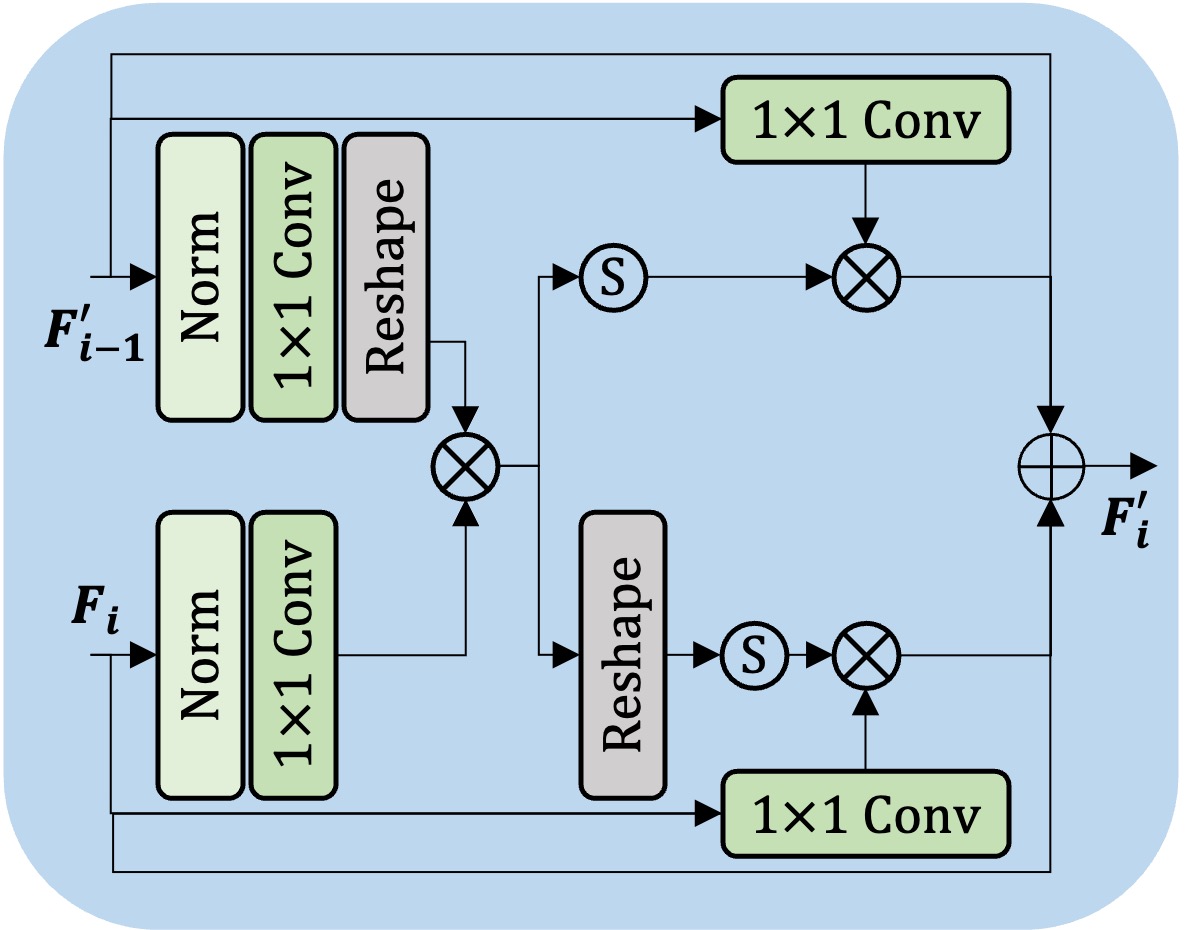}
    \caption{The structure of feature cross-attention mechanism (FCAM).}
    \label{fig:fam}
\end{figure}

As illustrated in Figure~\ref{fig:fam}, for simplicity, we take the example that the memory bank has only one storage size. The fusion process follows a similar approach to mainstream feature fusion methods, leveraging an attention mechanism to retain the most informative content through mutual comparison of features. The key distinction from existing methods~\cite{FSNet} lies in our objective: enhancing the local information within the current feature $F_i$. By integrating historical information from the memory bank into the current feature, the resulting representation not only preserves the global context captured by Mamba but also incorporates enriched local details. Specifically, given the current feature $F_i$ and historical information $F_{i-1}'$, we first obtain the $Q, K, V$ matrices used to perform the attention computation as follows:
\begin{equation}
\begin{aligned}
\label{equ:mem4}
    &Q_i = K_{i-1} = f_{1 \times 1}^c(Norm(F_i))
    \\
    &Q_{i-1} = K_i  = Reshape(f_{1 \times 1}^c(Norm(F_{i-1}')))
    \\
    & V_i = f_{1 \times 1}^c(F_i)
    \\
    & V_{i-1} = f_{1 \times 1}^c(F_{i-1}')
\end{aligned}
\end{equation}

Next up, we obtain the final enhanced feature $F_i'$ by performing the attention calculation and feature fusion as follows:
\begin{equation}
\begin{aligned}
\label{equ:mem5}
    &F_i' = Att_{i \rightarrow i-1} \oplus Att_{i-1 \rightarrow i} \oplus F_i \oplus F_{i-1}'
    \\
    &Att_{i \rightarrow i-1} = SoftMax(Reshape(Q_i \otimes K_i)) \otimes V_i
    \\
    & Att_{i-1 \rightarrow i} = SoftMax(Q_{i-1} \otimes K_{i-1}) \otimes V_{i-1}
\end{aligned}
\end{equation}

Finally, we concatenate all the features stored in the memory bank to obtain the final output feature of MemVSSM, denoted as $F_{out}$:
\begin{equation}
\label{equ:mem6}
F_{out} = concatenate(F_i', F_{i-1}' ,...,F_{i-K-1}')
\end{equation}

\begin{algorithm}[t]
\caption{Calculation of Ising Loss}
\label{alg:ising}
\begin{algorithmic}[1]
\REQUIRE Predicted image $\hat{I}$ with shape $C \times H \times W$
\STATE Initialize loss $\mathcal{L}_{\text{Ising}} = 0$
\FOR{each pixel $(x, y)$ in $\hat{I}$}
    \FOR{each neighboring pixel $(x', y')$ of $(x, y)$}
        \FOR{channel $c = 1$ to $C$}
            \STATE $\mathcal{L}_{\text{Ising}} \mathrel{+}= \left| \hat{I}_c(x, y) - \hat{I}_c(x', y') \right|$
        \ENDFOR
    \ENDFOR
\ENDFOR
\STATE Normalize: $\mathcal{L}_{\text{Ising}} \mathrel{/}= C \times H \times W$
\RETURN $\mathcal{L}_{\text{Ising}}$
\end{algorithmic}
\end{algorithm}

\subsection{Ising-inspired Regularization Loss}
Since the input is processed in chunks for Mamba, the final restoration results often suffer from a lack of spatial coherence when relying solely on  standard reconstruction loss, resulting in oversmoothed textures or fragmented structures. Inspired by the Ising model from statistical physics—which encourages local consistency among adjacent elements—we propose a novel Ising Loss to enhance spatial continuity and alleviate artifacts introduced by chunk-wise processing in Mamba-based architectures. This loss promotes similarity among neighboring pixel representations, thereby improving spatial coherence and enabling structure-aware image deblurring.

Specifically, given the predicted image $\hat{I}$, we define the Ising loss as the sum of absolute differences between a pixel and its neighboring pixels:

\begin{equation}
\label{equ:is}
\mathcal{L}_{\text{Ising}} = \frac{1}{Z} \sum_{c=1}^{C} \sum_{x=1}^{H} \sum_{y=1}^{W} \sum_{(x', y') \in \mathcal{N}(x, y)} \left| \hat{I}_c(x, y) - \hat{I}_c(x', y') \right|,
\end{equation}
where $\mathcal{N}(x, y)$ denotes the local neighborhood of pixel $(x, y)$ (e.g., four- or eight-connected neighbors), and $Z = C \times H \times W$ is a normalization factor to make the loss invariant to image size.

The computation of $\mathcal{L}_{\text{Ising}}$ is outlined in Algorithm~\ref{alg:ising}. At each pixel location, we compute the sum of absolute differences with its neighbors and accumulate the result over all channels and spatial locations. Finally, the aggregated loss is normalized.
To form the total training objective, we combine the Ising loss with standard reconstruction loss:

\begin{equation}
\begin{aligned}
\label{eq:loss1}
\mathcal{L} &= \mathcal{L}_{c}(\hat{I},\overline I)  + \delta \mathcal{L}_{e}(\hat{I},\overline I) + \lambda \mathcal{L}_{f}(\hat{I},\overline I)) + \mathcal{L}_{\text{Ising}}
\\
\mathcal{L}_{c} &= \sqrt{||\hat{I} -\overline I||^2 + \epsilon^2}
\\
\mathcal{L}_{e} &= \sqrt{||\triangle \hat{I} - \triangle \overline I||^2 + \epsilon^2}
\\
\mathcal{L}_{f} &= ||\mathcal{F}(\hat{I})-\mathcal{F}(\overline I)||_1
\end{aligned}
\end{equation}
where  $\overline I$ denotes the target images and $\mathcal{L}_{c}$ is the  Charbonnier loss with constant $\epsilon = 0.001$. $\mathcal{L}_{e}$ is the edge loss, where $\triangle$ represents the  Laplacian operator. $\mathcal{L}_{f}$  denotes the frequency domains loss, and $\mathcal{F}$ represents fast Fourier transform. To control the relative importance of loss terms, we set the parameters $\lambda = 0.1$ and $\delta = 0.05$  as in~\cite{Zamir2021MPRNet,FSNet}.

\section{Experiments}
We begin by detailing the experimental setup of the proposed MBMamba. Next, we provide both qualitative and quantitative comparisons with state-of-the-art methods. This is followed by ablation studies to verify the effectiveness of our design. Lastly, we evaluate the resource efficiency of MBMamba.  The best and second best scores are \textbf{highlighted} and \underline{underlined}.

\subsection{Experimental Settings}
\subsubsection{Datasets}
In line with recent approaches~\cite{FSNet,Zamir2021MPRNet}, we train MBMamba on the GoPro dataset~\cite{Gopro}, which consists of 2,103 training image pairs and 1,111 evaluation pairs. To evaluate the model's generalization capability, we directly test the GoPro-trained model on the HIDE~\cite{HIDE} and RealBlur~\cite{realblurrim_2020_ECCV} datasets. The HIDE dataset, designed for human-centric motion deblurring, comprises 2,025 images. While both GoPro and HIDE are synthetically generated, RealBlur contains real-world image pairs, divided into two subsets: RealBlur-J and RealBlur-R.

\subsubsection{Training details}
We adopt the Adam optimizer~\cite{2014Adam} with $\beta_1 = 0.9$ and $\beta_2 = 0.999$. The learning rate is initially set to $5 \times 10^{-4}$ and decayed to $1 \times 10^{-7}$ using a cosine annealing schedule~\cite{2016SGDR}. Training is conducted on $256 \times 256$ image patches with a batch size of 32 over $4 \times 10^5$ iterations. Data augmentation is performed through horizontal and vertical flipping. Furthermore, we construct three variants of MBMamba by adjusting the number of sub-decoders $n$ (as shown in Figure~\ref{fig:network}(a)): MBMamba-S with 1 sub-decoder, MBMamba-B with 2 sub-decoders, and MBMamba-L with 4 sub-decoders.

\begin{table}[ht]
\centering
\caption{Quantitative evaluations of the proposed approach against state-of-the-art motion deblurring methods. \label{tb:deblurgh}}
\resizebox{\linewidth}{!}{
\begin{tabular}{ccccc}
    \hline
    \multicolumn{1}{c}{} & \multicolumn{2}{c}{GoPro}  & \multicolumn{2}{c}{HIDE} 
    \\
   Methods & PSNR $\uparrow$ & SSIM $\uparrow$ & PSNR $\uparrow$ & SSIM $\uparrow$   
    \\
    \hline\hline
    MPRNet~\cite{Zamir2021MPRNet} & 32.66 & 0.959 & 30.96 & 0.939 
    \\
    NAFNet-64~\cite{chen2022simple}&33.62&0.967&-&-
    \\
    Restormer~\cite{Zamir2021Restormer} & 32.92 & 0.961 & 31.22 & 0.942 
    \\
    FSNet~\cite{FSNet} &33.29&0.963 &31.05 & 0.941 
    \\
     DeblurDiNAT-L~\cite{DeblurDiNAT}&33.42 &0.965 & 31.28 &0.943
     \\
     MambaIR~\cite{guo2024mambair}&33.21 &0.962 &31.01 &0.939
    \\
    ALGNet~\cite{algnetgao2024learning}& 34.05 &\underline{0.969} &31.68 &\underline{0.952}
    \\
    LoFormer~\cite{xintm2024LoFormer} & 34.09 & \underline{0.969} & 31.86 &0.949
    \\
    XYScanNet~\cite{liu2024xyscannet} &33.91 &0.968 & 31.74 &0.947
    \\
    PGDN~\cite{PGDNFang_2025_CVPR}& 34.17 &0.950 & - & -
    \\
    MDT~\cite{MDTChen_2025_CVPR}& 34.26&\underline{0.969} &31.84 &0.948
    \\
    ACL~\cite{ACLGu_2025_CVPR} &33.25 &0.964 &- &-
    \\
    Omni-Deblurring~\cite{10919160}&33.29 &0.963& 31.65 &0.947
    \\
    MambaIRv2~\cite{guo2025mambairv2}&33.62 &0.967 &31.63 &0.948
    \\
    \hline
    \textbf{MBMamba-S(Ours)} &33.89 &0.968 &31.72 &0.947
    \\
    \textbf{MBMamba-B(Ours)} &\underline{34.33} &\underline{0.969}	&\underline{31.89}	&0.949
    \\
    \textbf{MBMamba-L(Ours)} &\textbf{34.68} & \textbf{0.972} & \textbf{32.22} & \textbf{0.953}
    \\
    \hline
\end{tabular}}
\end{table}

\begin{figure*} % use float package if you want it here
	\centering
	\includegraphics[width=1\linewidth]{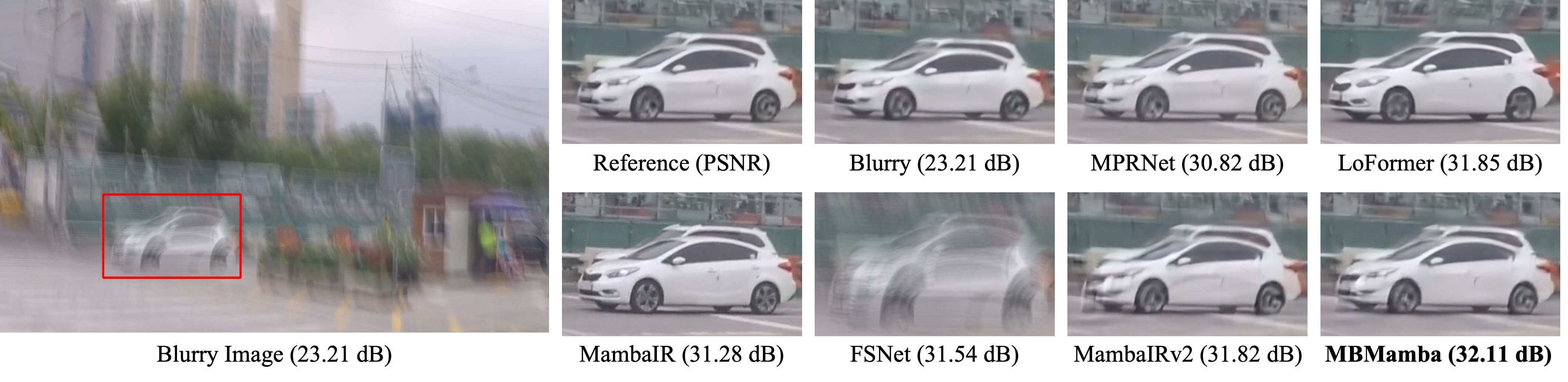}
	\caption{Image deblurring comparisons on the synthetic dataset~\cite{Gopro}. }
	\label{fig:blurm}
\end{figure*}

\subsection{Experimental Results}
\subsubsection{ Evaluations on the synthetic dataset }
Tables~\ref{tb:deblurgh} present the performance of various image deblurring methods on the synthetic GoPro~\cite{Gopro} and HIDE~\cite{HIDE} datasets. Compared to the previous state-of-the-art MDT~\cite{MDTChen_2025_CVPR}, our MBMamba-L achieves a 0.42 dB gain on the GoPro dataset. Against other Mamba-based approaches, MBMamba-S surpasses MambaIRv2~\cite{guo2025mambairv2} by 0.27 dB, while MBMamba-L delivers a notable improvement of 1.06 dB. Furthermore, relative to the best-performing Mamba variant LoFormer~\cite{xintm2024LoFormer}, MBMamba-L yields a significant enhancement of 0.57 dB. As shown in Figure~\ref{fig:param}, MBMamba’s performance scales favorably with increased model size, highlighting its strong scalability. Although trained exclusively on the GoPro dataset, our model achieves a 0.36 dB PSNR improvement over LoFormer on the HIDE dataset, demonstrating strong generalization capability. Visual comparisons in Figure~\ref{fig:blurm} further confirm that our method produces more visually appealing results.

\begin{table}
\centering
\caption{Quantitative real-world deblurring results.}
\label{tb:0deblurringreal}
\resizebox{\linewidth}{!}{
\begin{tabular}{ccccc}
    \hline
    \multicolumn{1}{c}{} & \multicolumn{2}{c}{RealBlur-R}  & \multicolumn{2}{c}{RealBlur-J} 
    \\
   Methods & PSNR $\uparrow$ & SSIM $\uparrow$ & PSNR $\uparrow$ & SSIM $\uparrow$   
    \\
    \hline
    \hline
DeblurGAN-v2~\cite{deganv2} & 36.44 & 0.935& 29.69& 0.870
\\
    MPRNet~\cite{Zamir2021MPRNet} & 39.31 & 0.972 & 31.76 & 0.922
   \\
Stripformer~\cite{Tsai2022Stripformer} & 39.84 & 0.975 & 32.48 & 0.929
\\
FFTformer~\cite{kong2023efficient}&40.11& 0.973 &32.62 &0.932
\\
MRLPFNet~\cite{MRLPFNet}& 40.92 &0.975 &33.19 &0.936
\\
 MambaIR~\cite{guo2024mambair}& 39.92 & 0.972 & 32.44 & 0.928
\\
ALGNet~\cite{algnetgao2024learning}&\underline{41.16}&\textbf{0.981}&32.94&0.946
    \\

     LoFormer~\cite{xintm2024LoFormer}&40.23 &0.974 &32.90 &0.933
    \\
    MambaIRv2~\cite{guo2025mambairv2}&40.36 &0.979 &32.91 &0.940
    \\
 \hline
    \textbf{MBMamba-S(Ours)}& 41.08 & 0.977 & 32.88 & 0.946
    \\
     \textbf{MBMamba-B(Ours)}& 41.12 & \underline{0.980} & \underline{33.23} & \underline{0.951}
    \\
      \textbf{MBMamba-L(Ours)}& \textbf{41.21} & \textbf{0.981} & \textbf{33.41} & \textbf{0.953}
    \\
    \hline
\end{tabular}}
\end{table}

\begin{figure*} % use float package if you want it here
	\centering
	\includegraphics[width=1\linewidth]{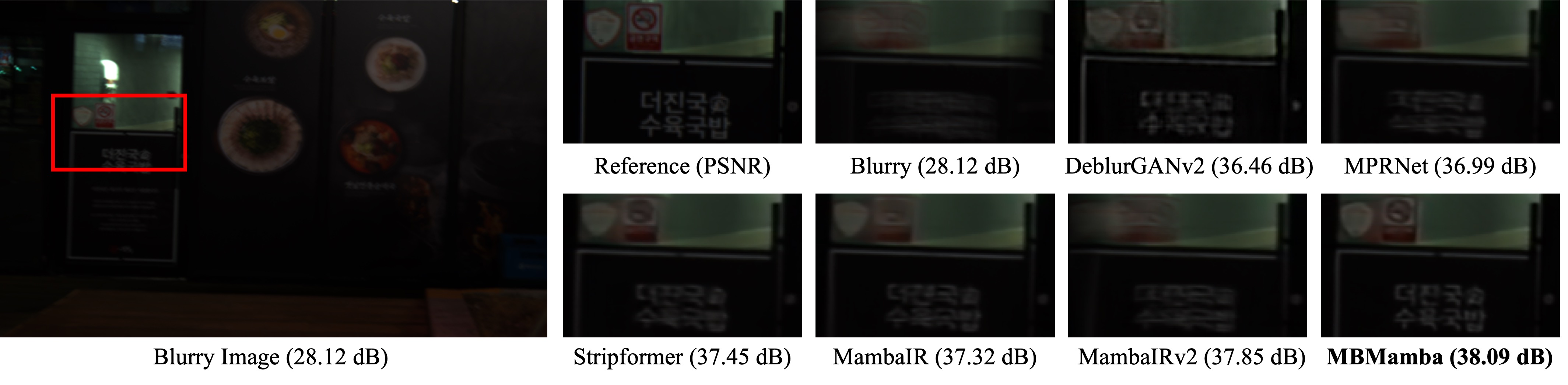}
	\caption{Image deblurring comparisons on the real-world  dataset~\cite{realblurrim_2020_ECCV}. }
	\label{fig:blurmreal}
\end{figure*}

\subsubsection{ Evaluations on the real-world dataset}
We further evaluate MBMamba on real-world images from the RealBlur dataset~\cite{realblurrim_2020_ECCV}. As shown in Table~\ref{tb:0deblurringreal}, our method achieves higher PSNR and SSIM scores than previous approaches. In particular, while the improvement over the prior best method, ALGNet~\cite{algnetgao2024learning}, is modest on the RealBlur-R dataset, it is more pronounced on the RealBlur-J dataset, with a PSNR gain of 0.47 dB. Visual comparisons in Figure~\ref{fig:blurmreal} further demonstrate that our model produces images with sharper details and better resemblance to the ground truth than competing methods.

\begin{table}
    \centering
    \caption{Ablation study on individual components of the
proposed MBMamba.}
    \label{tab:abl1}
    \begin{tabular}{cc}
    \hline
         Method&  PSNR
         \\
         \hline
         Baseline & 32.83
         \\
        Baseline replace with MemVSSM& 33.64
        \\
          Baseline + Ising loss & 32.85
         \\
         Baseline replace with MemVSSM + Ising loss & 33.89
         \\
         \hline
    \end{tabular}
\end{table}

\subsection{Ablation Studies}
We perform ablation studies to assess the effectiveness and scalability of our proposed method on the GoPro dataset~\cite{Gopro}, with the results summarized in Table~\ref{tab:abl1}. Using NAFNet~\cite{chen2022simple} as the baseline, we gradually introduce our proposed modules to analyze their individual and combined contributions. As shown in Table~\ref{tab:abl1}, replacing the baseline with MemVSSM significantly enhances the model’s ability to capture global information, leading to a notable improvement of approximately 0.81 dB. Additionally, introducing the Ising loss alone on the CNN-based baseline yields minimal performance gain, as the baseline already effectively models local information and benefits less from the smoothness regularization provided by the Ising loss. However, when combined with MemVSSM, the use of Ising loss noticeably boosts the model’s sensitivity to local structures, resulting in a further performance increase of about 0.25 dB. These results demonstrate the strong complementarity between our proposed modules.

\begin{table}
    \centering
       \caption{Impact of MemVSSM design choices on model performance.}
    \label{tab:ptsabl}

    \begin{tabular}{ccc}
    \hline
         Net& PSNR & $\triangle$ PSNR  
         \\
         \hline\hline
          VSSM~\cite{guo2024mambair}& 33.37 &-  
         \\
         ASSM~\cite{guo2025mambairv2}& 33.65  &+0.28 
         \\
         \textbf{MemVSSM(Ours)} & 33.89& +0.52
         \\
         \hline
    \end{tabular}
\end{table}

\begin{figure}[htb] % use float package if you want it here
	\centering
	\includegraphics[width=1\linewidth]{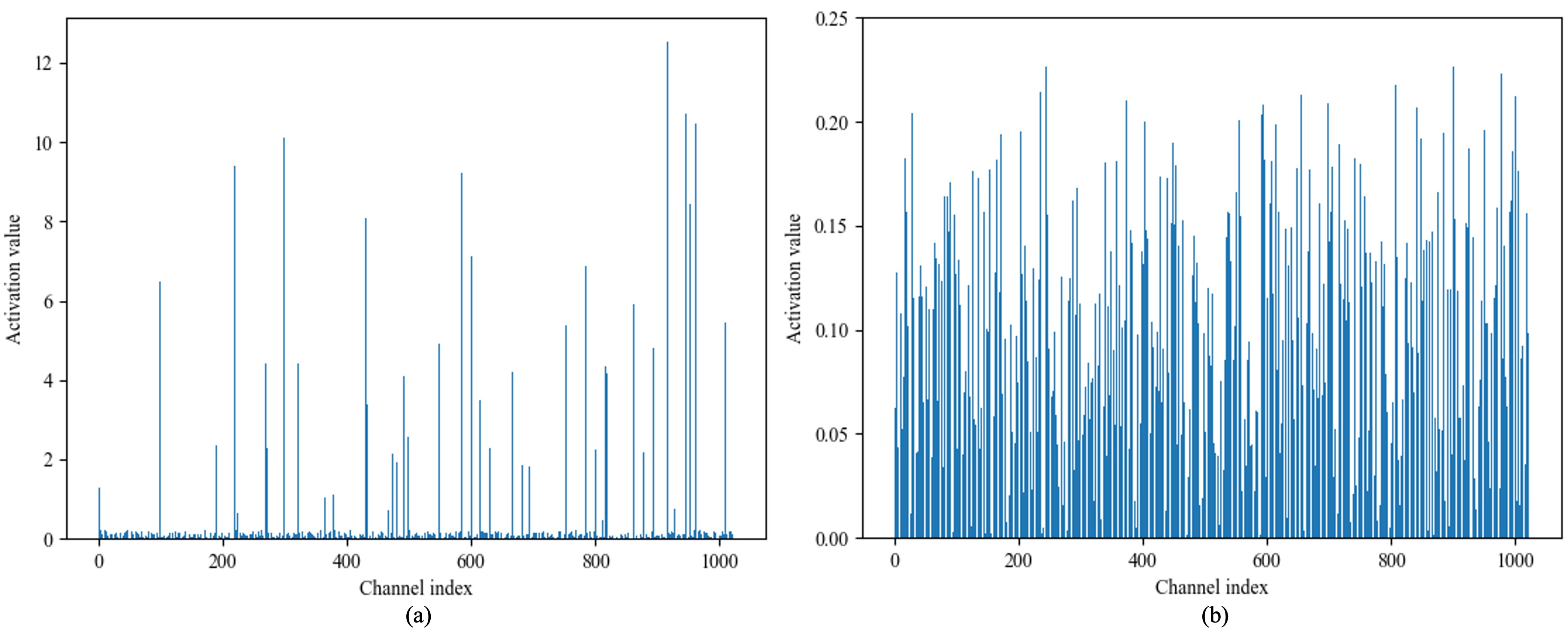}
	\caption{Following~\cite{guo2024mambair}, we use ReLU activation followed by global average pooling to compute channel activation values. (a) For the standard SSM, a significant portion of channels remain inactive, revealing channel redundancy. (b)The outputs from our MemVSSM module.}
	\label{fig:chann}
\end{figure}

To further assess the effectiveness of our MemVSSM module, we replace it with existing VSSM methods~\cite{guo2024mambair,guo2025mambairv2}. As shown in Table~\ref{tab:ptsabl}, our MemVSSM achieves the best performance among all variants. Additionally, we apply ReLU activation followed by global average pooling on the MemVSSM outputs to compute channel activation values (see Figure~\ref{fig:chann}). The results clearly indicate that MemVSSM effectively mitigates the problem of channel redundancy caused by an excessive number of hidden states in the state space model.

Since our MBMamba employs an existing pre-trained model as the encoder, we assess the influence of the pre-trained weights on overall performance. As shown in Table~\ref{tab:pretrain}, leveraging pre-trained models clearly enhances performance. Furthermore, we investigate the effect of freezing encoder parameters and find that it has minimal impact on the results. Therefore, to conserve computational resources, we choose to freeze the encoder and train only the decoder.

\begin{table}
    \centering
       \caption{Effect of the pre-trained models.}
    \label{tab:pretrain}
     \resizebox{\linewidth}{!}{
    \begin{tabular}{ccccc}
    \hline
         Net&Pre-trained & Trainable&PSNR & $\triangle$ PSNR
         \\
         \hline\hline
         (a)&     & \ding{52} & 33.74 & -
         \\
         (b)&\ding{52}    &\ding{52}   &33.86 & +0.12
         \\
         (c) &\ding{52}   &   &33.89 & +0.15
         \\
         \hline
    \end{tabular}}
\end{table}

\subsection{Resource Efficient}
To further demonstrate the resource efficiency of our approach, we evaluate the model complexity and compare it with state-of-the-art methods in terms of inference time and FLOPs. Although our MemVSSM processes input features sequentially—thus not fully leveraging the parallelism inherent in the original SSM—it nonetheless achieves impressive efficiency. As shown in Table~\ref{tab:computational} and Figure~\ref{fig:param}, our MBMamba model not only delivers state-of-the-art performance but also substantially reduces computational overhead.
Specifically, MBMamba-L surpasses the previous best-performing method, MambaIRv2~\cite{guo2025mambairv2}, by 1.06 dB, while cutting computational cost by up to 84.2\% and achieving nearly 1.5× faster inference. These results clearly demonstrate the effectiveness and efficiency of our method in balancing performance with resource usage.

\begin{table}
    \centering
    \caption{The evaluation of model computational complexity on the GoPro dataset~\cite{Gopro}.}
    \label{tab:computational}
    \resizebox{\linewidth}{!}{
    \begin{tabular}{ccccc}
    \hline
         Method& Time(s) & FLOPs(G)  & PSNR & SSIM
         \\
         \hline\hline
         MPRNet~\cite{Zamir2021MPRNet} & 1.148 & 777 & 32.66 &0.959
         \\
         Restormer~\cite{Zamir2021Restormer} & 1.218 & 140 & 32.92 & 0.961
         \\
         IRNeXt~\cite{IRNeXt} &\underline{0.255} & 114 & 33.16 & 0.962
         \\
         FSNet~\cite{FSNet} &0.362 & 111& 33.29&0.963
         \\
         MambaIR~\cite{guo2024mambair} & 0.743 & 439 & 33.21 & 0.962
         \\
          MambaIRv2~\cite{guo2025mambairv2} & 0.743 & 664 & 33.62 & 0.967
         \\
         \hline
         \textbf{MBMamba-S(Ours)} &\textbf{0.249} &\textbf{54} & 33.89 & 0.968
         \\
           \textbf{MBMamba-B(Ours)} &0.283 &\underline{74} & \underline{34.33} & \underline{0.969}
         \\
           \textbf{MBMamba-L(Ours)} &0.442&105 & \textbf{34.68} & \textbf{0.972}
         \\
         \hline
    \end{tabular}}
\end{table}

\section{Conclusion}
In this paper, we present a structure-aware image deblurring network that effectively integrates both local and global features, without incurring the overhead of multiple scans or added modules. Specifically, we design a memory buffer mechanism to store and reuse historical information, facilitating more reliable modeling of the relevance between adjacent features. In addition, we introduce an Ising-inspired regularization loss inspired by physical systems, which simulates the "mutual attraction" between neighboring pixels through energy minimization. This loss encourages structural consistency and promotes smoother, more coherent image restoration. Extensive experimental results demonstrate that our proposed method achieves superior performance compared to state-of-the-art approaches.

\bibliographystyle{IEEEtran}
\bibliography{aaai2026}

\section{Appendix}
\label{app}

\subsection{Overview}

Mathematical Interpretation of MemVSSM~\ref{sec:math}

Dataset~\ref{sec:data}

More Ablation Studies~\ref{sec:mas}

% Resource Efficient~\ref{sec:re}

Additional Visual Results~\ref{sec:Visual}

\subsection{Mathematical Interpretation of MemVSSM}
\label{sec:math}

Consider one MemVSSM processing channel-chunk \(N\). Let $ x^{(n)}\in\mathbb{R}^{B\times d\times H\times W}$
be the input chunk and \(T=H\cdot W\) the number of spatial tokens per chunk. Let $M(\cdot)$ denote the Mamba mapping  applied to the chunk tokens, and define the Mamba output
\[
o^{(n)} = M\big(x^{(n)}\big)\in\mathbb{R}^{B\times d\times H\times W}.
\]
The implementation stores a fused memory \(m^{(k)}\) after processing chunk \(k\) as
\[
m^{(k)} := \operatorname{detach}\big(y^{(n)}\big),
\]
where \(y^{(n)}\) is the fused output produced for chunk \(n\) (see below); `detach` indicates stopping gradients.

We also let \(C\) be the total channels and \(N\) the number of chunks so that the per-chunk channel dimension is
\[
d = \frac{C}{N}.
\]
We denote tokenized representations by indices \(i,j\in\{1,\dots,T\}\). FCAM computes scaled dot-product attention between current chunk queries (from \(o^{(n)}\)) and memory keys (from \(m^{(k-1)}\)):
\begin{align}
a_{ij} &= \frac{Q_\ell[i]\cdot Q_m[j]}{\sqrt{d}}, \\
S_{ij} &= \operatorname{softmax}_j(a_{ij}),\qquad \sum_j S_{ij}=1,
\end{align}
and two attention-propagated contributions (written per-token and then reshaped):
\begin{align}
G_\ell[i] &= \beta \sum_{j=1}^{T} S_{ij}\, V_m[j], \label{eq:G_l}\\
G_m[j]   &= \gamma \sum_{i=1}^{T} S'_{ji}\, V_\ell[i], \label{eq:G_m}
\end{align}
where \(S'=\operatorname{softmax}(a^\top)\), and \(\beta,\gamma\in\mathbb{R}^{d}\) are the per-channel learnable scales (initialized at zero in code). The fused forward output for chunk \(n\) is therefore
\begin{equation}\label{eq:fused_output}
y^{(n)} \;=\; o^{(n)} \;+\; G_\ell \;+\; m^{(k-1)} \;+\; G_m,
\end{equation}
with the implementation detail that \(m^{(k-1)}=\operatorname{detach}(y^{(n-1)})\).

The baseline (no fusion) per-chunk update is \(y^{(n)}_{\text{base}} = x^{(n)} + o^{(n)} + \text{(FFN residual)}\). With fusion, the Mamba output \(o^{(n)}\) is additively augmented by attention-weighted transforms of the previous chunk's memory and by a direct additive memory term (Eq.~\ref{eq:fused_output}). Thus fusion changes the forward dynamics seen by subsequent model components: information from chunk \(n-1\) influences the activations of chunk \(n\) both through attention (\(G_\ell\)) and direct addition (\(m^{(k-1)}\)). Concretely, fusion yields richer cross-channel / cross-spatial coupling in the activations received by downstream layers.

Because the implementation writes memory via \(\ m^{(k-1)}=\operatorname{detach}(y^{(n-1)})\), the gradient path through memory is blocked. Using chain-rule notation,
\[
\frac{\partial L}{\partial y^{(n-1)}} \;\not\ni\; \frac{\partial y^{(n)}}{\partial m^{(k-1)}} \frac{\partial L}{\partial y^{(n)}},
\]
since \(\partial m^{(k-1)}/\partial y^{(n-1)}=0\). 
Therefore, although later chunks receive activations influenced by earlier chunks, the gradients do not accumulate recursively through the memory. This prevents the long-horizon gradient explosion or saturation typically observed in fully differentiable recurrent state updates. Gradients from the loss at chunk $n$ still update the FCAM parameters (including the projection layers and $\beta$, $\gamma$), but they do not backpropagate into the Mamba parameters that generated $m^{(k-1)}$. In this design, the fusion modifies the forward dynamics by strengthening cross-chunk coupling in the activations, yet it keeps the backward dynamics local to each chunk. This locality is the key reason the implementation avoids gradient explosion or saturation when processing many chunks.

Because \(S_{ij}\) is a row-normalized softmax,
\begin{equation}\label{eq:bound}
\|G_\ell[i]\| \le \|\beta\|_\infty \sum_j S_{ij}\|V_m[j]\| \le \|\beta\|_\infty \max_{j}\|V_m[j]\|.
\end{equation}
Since \(\beta\) and \(\gamma\) are initialized to zero and learned gradually, the early-stage contribution of fusion is small, providing numerical stability. Combined with the softmax normalization, Eqs.~\ref{eq:G_l}--\ref{eq:bound} show fusion cannot produce unbounded outputs unless the learned scales diverge.

The memory fusion in MemVSSM  effectively enhances Mamba with local historical context while preserving stable gradient flow. The behavior of the mechanism depends smoothly on $K$ and $N$ offering a controllable trade-off between local detail reinforcement and computational cost.

\subsection{Datasets}
\label{sec:data}
In line with recent approaches~\cite{FSNet,Zamir2021MPRNet}, we train MBMamba on the GoPro dataset~\cite{Gopro}, which consists of 2,103 training image pairs and 1,111 evaluation pairs. To evaluate the model's generalization capability, we directly test the GoPro-trained model on the HIDE~\cite{HIDE} and RealBlur~\cite{realblurrim_2020_ECCV} datasets. The HIDE dataset, designed for human-centric motion deblurring, comprises 2,025 images. While both GoPro and HIDE are synthetically generated, RealBlur contains real-world image pairs, divided into two subsets: RealBlur-J and RealBlur-R.

\begin{figure*} % use float package if you want it here
	\centering
	\includegraphics[width=1\linewidth]{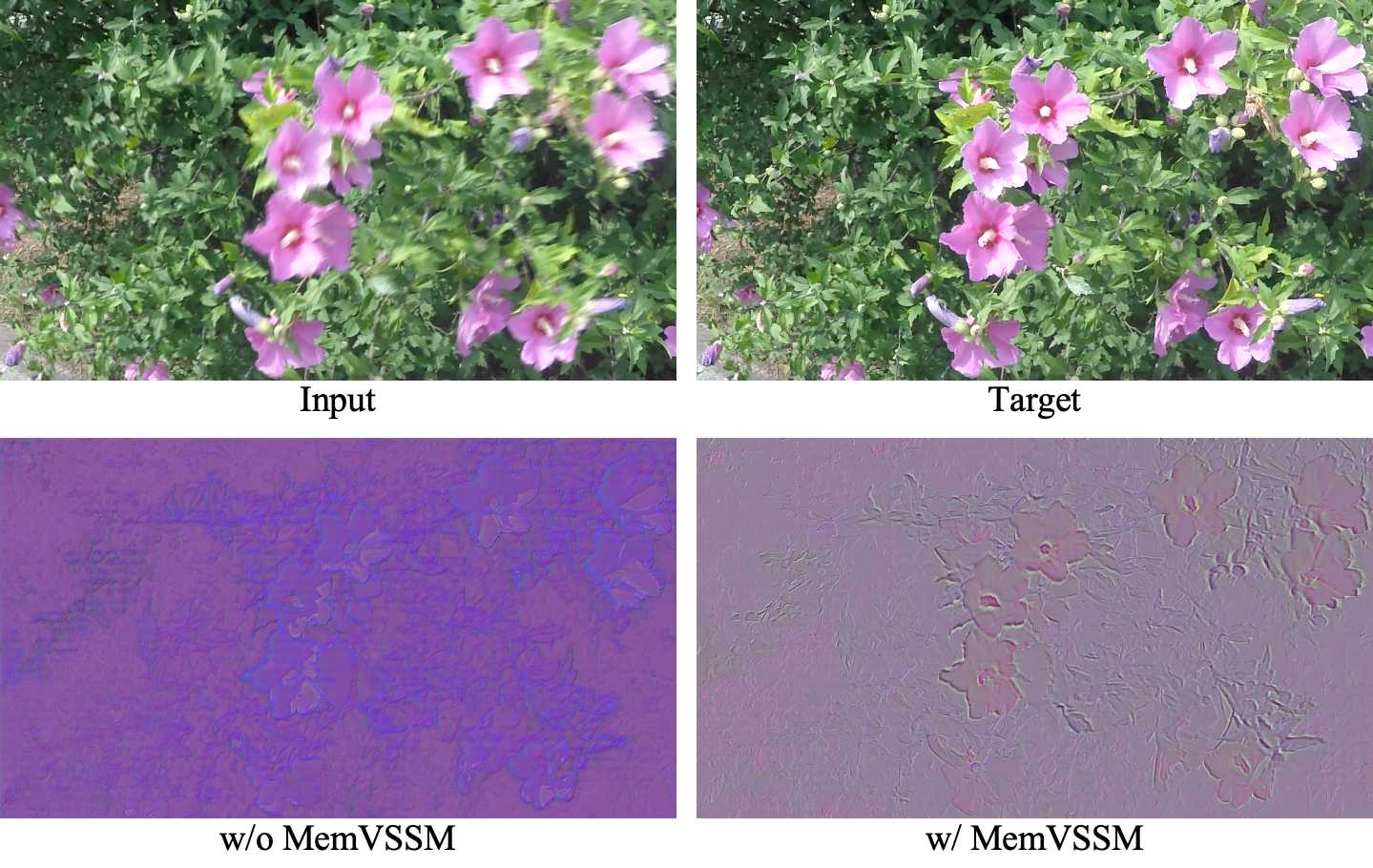}
	\caption{Effect of MemVSSM.}
	\label{fig:memvssm}
\end{figure*}

\begin{table}
    \centering
    \caption{The  impact of hyperparameters.}
    \label{tab:ablcmw}
    \begin{tabular}{ccccc}
    \hline
         $N$& $K$ & $\kappa$  & PSNR & $\triangle$ PSNR
         \\
         \hline\hline
         4 & 4 & 0.001 & 33.89 & -
         \\
         4 & 4 & 0.005 & 33.75 & -0.14
         \\
         2 & 2 & 0.001 & 33.53 & -0.36
         \\
         6 & 4 & 0.001 & 33.77 & -0.12
         \\
         6 & 4 & 0.005 & 33.86 & -0.03
         \\
         4 & 2 & 0.001 & 33.87 & -0.02 
         \\
         \hline
    \end{tabular}
\end{table}

\subsection{More Ablation Studies}
\label{sec:mas}

We present the visualization of feature maps in Figure~\ref{fig:memvssm} to emphasize the advantages of our proposed MemVSSM. As shown, the results clearly indicate that the features extracted with MemVSSM capture finer details and richer structural information compared to the baseline.

In addition to evaluating each hyperparameter individually, we analyze the interactions between the number of chunks $N$, memory bank depth $K$, and Ising loss weight $\kappa$. As shown in Table~\ref{tab:ablcmw}, chunk number $N$ has the most significant influence, but the combination of $N$, $K$, and $\kappa$ jointly determines optimal performance.  For larger chunk numbers ($N=6$), increasing $\kappa$ slightly improves PSNR (33.81 $\rightarrow$ 33.86). This suggests that the optimal Ising loss weight depends on $N$. Larger $N$ benefits from a slightly higher $\kappa$, while smaller $N$ prefers a lower $\kappa$ to avoid over-regularization. Since we only need the previous element in the memory bank to be fused with the current element, its depth does not have a big impact on the model performance. To facilitate the subsequent MemVSSM output, we set its depth to the same size as the chunk size.

\begin{table}
    \centering
    \caption{The  impact of Ising loss.}
    \label{tab:ablisloss}
    \begin{tabular}{ccc}
    \hline
        Loss&   PSNR & $\triangle$ PSNR
         \\
         \hline\hline
        Ising loss & 33.89 & -
        \\
        Laplacian loss & 33.74 & -0.15
        \\
         \hline
    \end{tabular}
\end{table}

\begin{figure*} % use float package if you want it here
	\centering
	\includegraphics[width=1\linewidth]{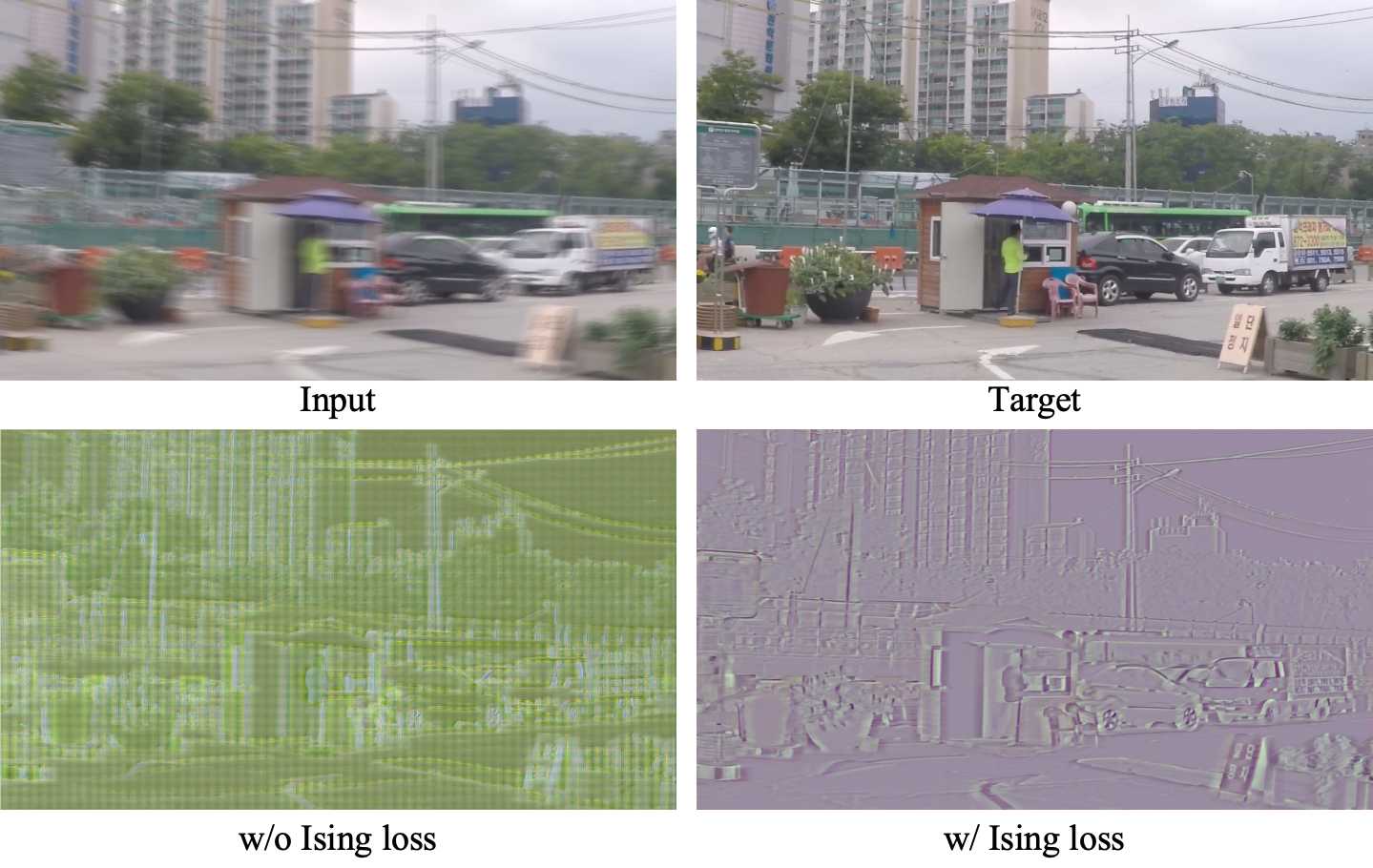}
	\caption{Effect of Ising loss.}
	\label{fig:isingloss}
\end{figure*}

To further demonstrate the effectiveness of Ising loss, we visualize its feature maps in Figure~\ref{fig:isingloss}. In our MemVSSM, inputs are processed in chunks, and relying solely on standard reconstruction loss often leads to oversmoothed textures and fragmented structures due to poor spatial coherence. Ising loss enhances spatial continuity, alleviates artifacts caused by chunk-wise processing, and promotes similarity among neighboring pixel representations, thereby improving spatial coherence and enabling structure-aware image deblurring.

We also replace the  Ising loss with Laplacian regularization. While Laplacian loss penalize local intensity differences uniformly to encourage smoothness, it tend to oversmooth fine structures and edges, which can degrade perceptual quality in image restoration tasks. In contrast, the Ising loss leverages a pairwise spin interaction model to selectively encourage piecewise-constant regions while preserving sharp boundaries. This leads to two distinct advantages: (i) enhanced preservation of high-frequency structures such as edges and textures, and (ii) stronger local consistency without introducing excessive blurring. Empirical results in Table~\ref{tab:ablisloss} confirm that Ising regularization achieves superior reconstruction quality.

\subsection{Additional Visual Results}
\label{sec:Visual}
\begin{figure*} % use float package if you want it here
	\centering
	\includegraphics[width=1\linewidth]{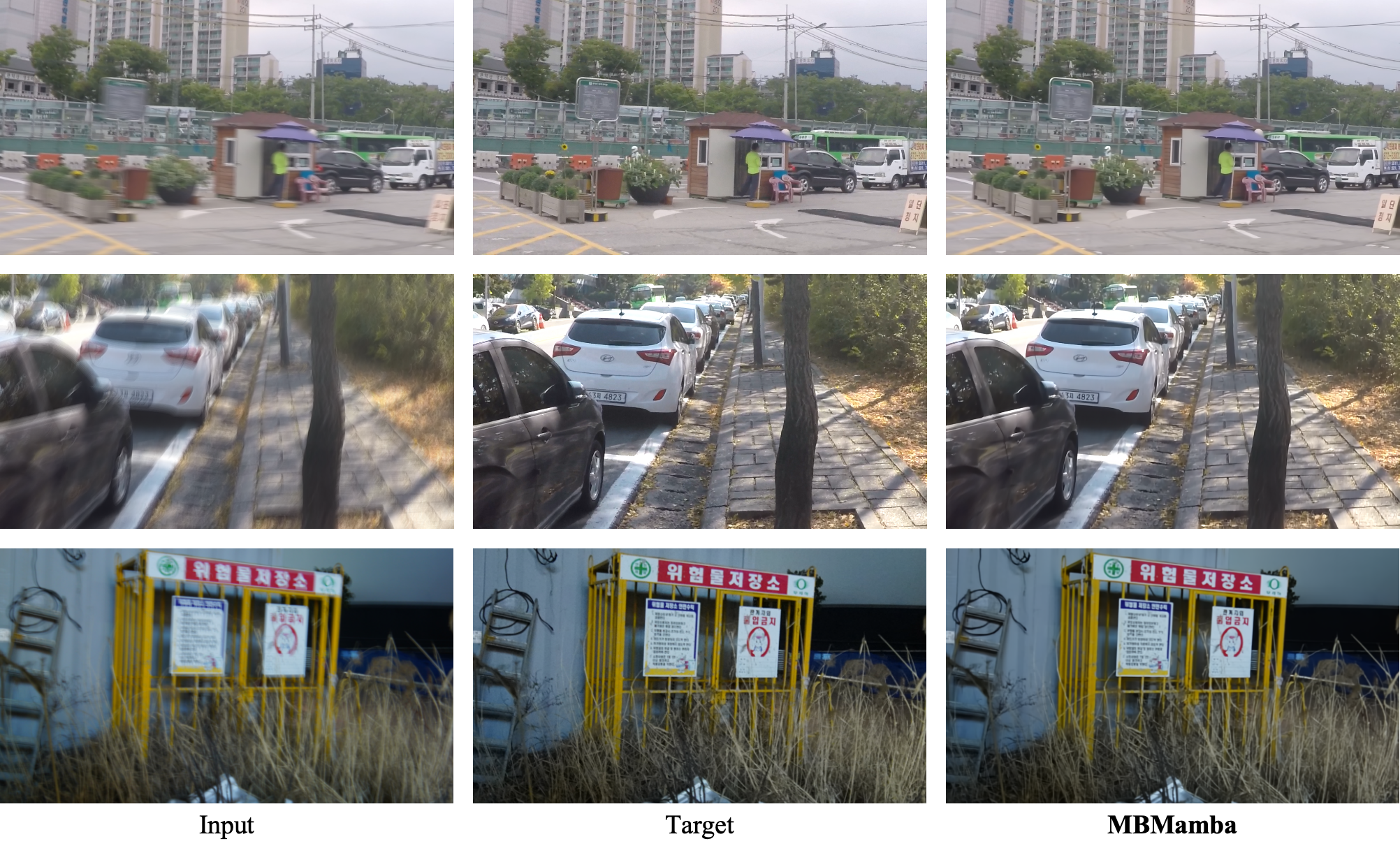}
	\caption{Visualization results on synthetic and real datasets.}
	\label{fig:mgopr}
\end{figure*}

In this section, we present additional visual results to highlight the effectiveness of our proposed approach, as shown in Figures~\ref{fig:mgopr}.
It is clear that our model produces more visually appealing outputs for both synthetic and real-world motion deblurring.

\vfill

\end{document}